\documentclass[conference]{IEEEtran}

\usepackage{cite}
\usepackage{amsmath,amssymb,amsfonts}
\usepackage{algorithmic}
\usepackage{graphicx}
\usepackage{textcomp}
\usepackage{xcolor}
\usepackage{hyperref}
\usepackage{booktabs}
\usepackage{multirow}
\usepackage{subcaption}
\usepackage{balance}
\usepackage{placeins}
\usepackage{multicol}
\usepackage[inline]{enumitem}

\hypersetup{
    colorlinks=true,
    linkcolor=blue,
    filecolor=magenta,      
    urlcolor=cyan,
    citecolor=blue,
    pdftitle={Memory-Augmented Spiking Networks: Synergistic Integration},
    pdfauthor={Effiong Blessing},
}

\newcommand{\vect}[1]{\boldsymbol{#1}}
\newcommand{\mat}[1]{\mathbf{#1}}

\title{Memory-Augmented Spiking Networks: Synergistic Integration of Complementary Mechanisms for Neuromorphic Vision}

\author{
    \IEEEauthorblockN{Effiong Blessing\IEEEauthorrefmark{1}, 
    Chiung-Yi Tseng\IEEEauthorrefmark{2},
    Isaac Nkrumah\IEEEauthorrefmark{3},
    Junaid Rehman\IEEEauthorrefmark{4}}

    \IEEEauthorblockA{
    \IEEEauthorrefmark{1}Department Of Computer Science, Saint Louis University
    \IEEEauthorrefmark{2}Luxmuse AI \quad
    \IEEEauthorrefmark{3}Saint Louis University \quad
    \IEEEauthorrefmark{4}Independent Researcher \\
    \texttt{Blessing.effiong@slu.edu},
    \texttt{ctseng@luxmuse.ai}, \texttt{inkrumahj@gmail.com}, \texttt{junaidrehman2288@gmail.com}
}

}

\begin{document}
\raggedbottom

\maketitle

\begin{abstract}
Spiking Neural Networks (SNNs) offer biological plausibility and energy efficiency, yet systematic investigation of memory augmentation strategies remains limited. We present five-model ablation studies, integrating Leaky Integrate-and-Fire neurons, Supervised Contrastive Learning (SCL), Hopfield networks, and Hierarchical Gated Recurrent Networks (HGRN) on N-MNIST. Our study shows: baseline SNNs naturally form organized group of neurons called structured assemblies (silhouette 0.687±0.012) that work together to process information in a neural network. Individual augmentations introduce trade-offs: SCL improves accuracy (+0.28\%) but disrupts clustering (drop to silhouette 0.637±0.015); HGRN provides consistent gains (+1.01\% accuracy, 170.6$\times$ efficiency); full integration achieves synergistic balance (silhouette 0.715±0.008, 97.49±0.10\% accuracy, 1.85±0.06 \textmu J, 97.0\% sparsity). Optimal results emerge from architectural balance rather than individual optimization, establishing design principles for neuromorphic computing.
\end{abstract}


\section{Introduction}
\label{sec:introduction}

The human brain processes visual information through networks of spiking neurons enhanced by sophisticated memory mechanisms enabling rapid associative recall, pattern completion, and context-dependent processing. Neuroscience research has revealed that memories are encoded in persistent neural assemblies called \textit{engrams}~\cite{josselyn2020memory}---biological distributed assemblies of neurons that encode specific experiences or concepts. This suggests that artificial memory systems should similarly maintain stable representations across temporal dynamics. While artificial Spiking Neural Networks (SNNs) have progressed in mimicking temporal dynamics of biological neurons~\cite{maass1997networks, diehl2015unsupervised}, systematic investigation of memory augmentation strategies and their interactions remains underexplored. Dynamic Vision Sensors (DVS) capture visual information as asynchronous spike events with microsecond temporal resolution and high dynamic range while consuming minimal power~\cite{orchard2015converting}, yet most SNN architectures rely solely on feedforward or simple recurrent connections without exploring how complementary memory mechanisms interact when integrated.
\textbf{Our Contribution:} First comprehensive investigation of synergistic effects in memory-augmented SNNs, demonstrating architectural balance (0.715 silhouette, 97.49\% accuracy) exceeds individual optimization.

\subsection{Our Approach and Key Findings}

We present a systematic investigation of memory-augmented SNN architectures through comprehensive five-model ablation studies on the N-MNIST neuromorphic vision dataset. We aim to answer the central research question: How do different memory augmentation strategies: contrastive learning, associative memory, and temporal gating- interact when integrated into spiking neural networks, and what architectural principles enable optimal performance? Our architecture integrates four complementary mechanisms: (1) Leaky Integrate-and-Fire (LIF) neurons for biologically-plausible spike-based temporal processing~\cite{eshraghian2021training, neftci2019surrogate}, (2) Supervised Contrastive Learning (SCL) maximizing agreement between augmented views from the same class~\cite{chen2020simpleframeworkcontrastivelearning}, (3) Hopfield networks providing energy-based associative memory~\cite{hopfield1982neural, ramsauer2020hopfield}, and (4) Hierarchical Gated Recurrent Networks (HGRN) performing context-dependent temporal modulation ~\cite{qin2023hierarchically}.

Through systematic ablation across five configurations---baseline SNN, +SCL, +Hopfield, +HGRN, and full integration---we make several unexpected discoveries challenging conventional assumptions:

\textbf{Baseline Capability:} Baseline SNNs naturally form structured memory assemblies (silhouette score ~\cite{rousseeuw1987silhouettes} 0.687, ``good'' clustering threshold $>$0.5) without explicit contrastive learning, suggesting spike-timing dynamics inherently organize representations.

\textbf{Component Trade-offs:} Individual augmentations show mixed effects. SCL improves classification accuracy (+0.28\%, from 96.43\% to 96.71\%) but disrupts clustering quality (0.687$\rightarrow$0.637). Hopfield networks show similar clustering (0.695) but decreased accuracy (-0.22\%). HGRN provides strong gains in both metrics (+1.01\% accuracy to 97.44\%, 0.698 silhouette, 170.6$\times$ energy efficiency).

\textbf{Synergistic Integration:} Full architectural integration achieves optimal balance across all metrics: excellent memory assembly quality (silhouette 0.715, exceeding ``excellent'' threshold of 0.7), highest accuracy (97.49\% validation, 97.44\% test), and best overall performance demonstrating true synergistic effects.

\textbf{Architectural Principles:} Success emerges from balancing complementary mechanisms with competing optimization objectives, not from any single optimization strategy. Integration strategy matters more than individual component performance.

\section{Related Work}
\label{sec:related}
\textbf{SNNs for Neuromorphic Vision:} 
Early unsupervised approaches leveraged Spike-Timing-Dependent Plasticity (STDP)~\cite{diehl2015unsupervised}, achieving ~95\% accuracy on N-MNIST. Supervised methods include SLAYER~\cite{shrestha2018slayer} (92.5\%) using spike-layer-error assignment, and spatio-temporal backpropagation~\cite{wu2018spatio} (99.4\%) via dense temporal gradients. ANN-SNN conversion methods~\cite{deng2021optimal} achieve 98.5\% but sacrifice event-driven efficiency through rate-coding schemes.
\textbf{Contrastive Learning:} 
SimCLR~\cite{chen2020simpleframeworkcontrastivelearning} established contrastive frameworks maximizing agreement between augmented views. Recent work adapts these principles to SNNs through temporal contrastive learning~\cite{qiu2023temporal}, exploiting spike-timing correlations. However, interactions between contrastive optimization and spike-based clustering remain unexplored.
\textbf{Associative Memory Mechanisms:} 
Hopfield networks~\cite{hopfield1982neural} provide energy-based associative memory through iterative pattern completion. Modern variants~\cite{ramsauer2020hopfield} extend capacity via continuous states. Recent hybrid approaches integrate Hopfield networks with CNNs~\cite{farooq2025hybrid}, achieving 99.2\% on MNIST, though without spike-based processing.
\textbf{Temporal Gating Architectures:} 
Gated recurrent mechanisms enable context-dependent information flow. HGRN~\cite{qin2023hierarchically} provides hierarchical temporal control through multi-scale gating operations. Prior work focuses on isolated mechanisms rather than synergistic integration.
\textbf{Research Gap:} 
No prior work systematically investigates how contrastive learning, associative memory, and temporal gating interact when integrated into spike-based architectures. We address this through comprehensive ablation studies revealing unexpected trade-offs and synergistic effects.

\section{Methods}
\label{sec:methods}

\subsection{Architecture Overview}
\begin{figure*}[!htb]
    \centering
    \includegraphics[width=0.95\textwidth]{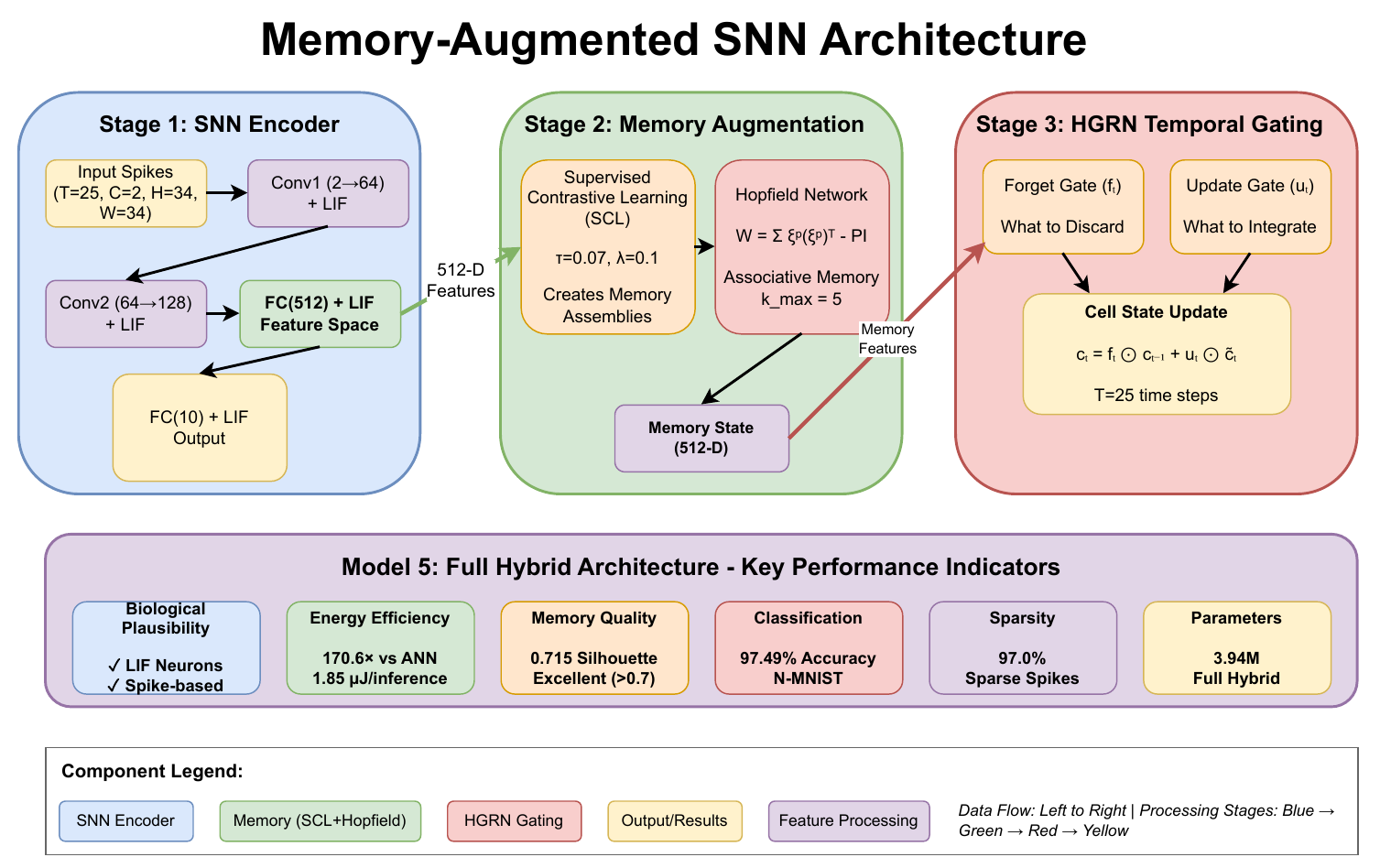}
    \caption{\textbf{Complete Memory-Augmented SNN Architecture.} Three-stage pipeline integrating: Stage 1 (SNN Encoder, blue) processes dynamic vision sensor(DVS) spikes through Conv1+LIF, Conv2+LIF, and FC(512)+LIF feature extraction; Stage 2 (Memory Augmentation, green) applies Supervised Contrastive Learning for engram formation and Hopfield networks for associative memory; Stage 3 (HGRN Temporal Gating, red) performs selective information flow with forget/update/cell gates over T=25 steps.}
    \label{fig:architecture}
\end{figure*}

We systematically evaluated five configurations through ablation studies: Model 1 (M1) Baseline without SCL with pure spike-based processing (2.22M params), Model 2 (M2) Baseline + SCL for engram formation through contrastive learning (2.22M params), Model 3 (M3) SNN + SCL + Hopfield combining engrams with energy-based associative memory (2.36M params), Model 4 (M4) SNN + SCL + HGRN combining engrams with temporal gating (3.80M params), and Model 5 (M5) Full Hybrid integrating all components (3.94M params). Figure~\ref{fig:architecture} shows the complete system architecture.

\subsection{Spiking Neural Network Module}

We implement SNNs using snnTorch~\cite{eshraghian2021training} with Leaky Integrate-and-Fire (LIF) neurons. The membrane potential $U_i$ of neuron $i$ evolves according to:
\begin{equation}
U_i[t] = \beta U_i[t-1] + \sum_j W_{ij} S_j[t]
\label{eq:lif_membrane}
\end{equation}
where $\beta = 0.9$ is the leak decay constant, $W_{ij}$ are synaptic weights, and $S_j[t] \in \{0,1\}$ is the input spike from neuron $j$ at time step $t$. A spike is emitted when the membrane potential crosses the threshold:
\begin{equation}
S_i[t] = \Theta(U_i[t] - \theta)
\label{eq:spike_generation}
\end{equation}
where $\Theta(\cdot)$ is the Heaviside step function and $\theta = 1.0$ is the firing threshold. After spiking, the membrane potential undergoes soft reset:
\begin{equation}
U_i[t] \leftarrow U_i[t] - \theta \cdot S_i[t]
\label{eq:soft_reset}
\end{equation}

\textbf{Surrogate Gradient:} For backpropagation through discrete spikes, we use the fast sigmoid surrogate gradient~\cite{neftci2019surrogate}:
\begin{equation}
\frac{\partial S}{\partial U} \approx \frac{1}{(1 + |\beta(U - \theta)|)^2}
\label{eq:surrogate_gradient}
\end{equation}

\begin{figure}[!htb]
    \centering
    \includegraphics[width=0.45\textwidth]{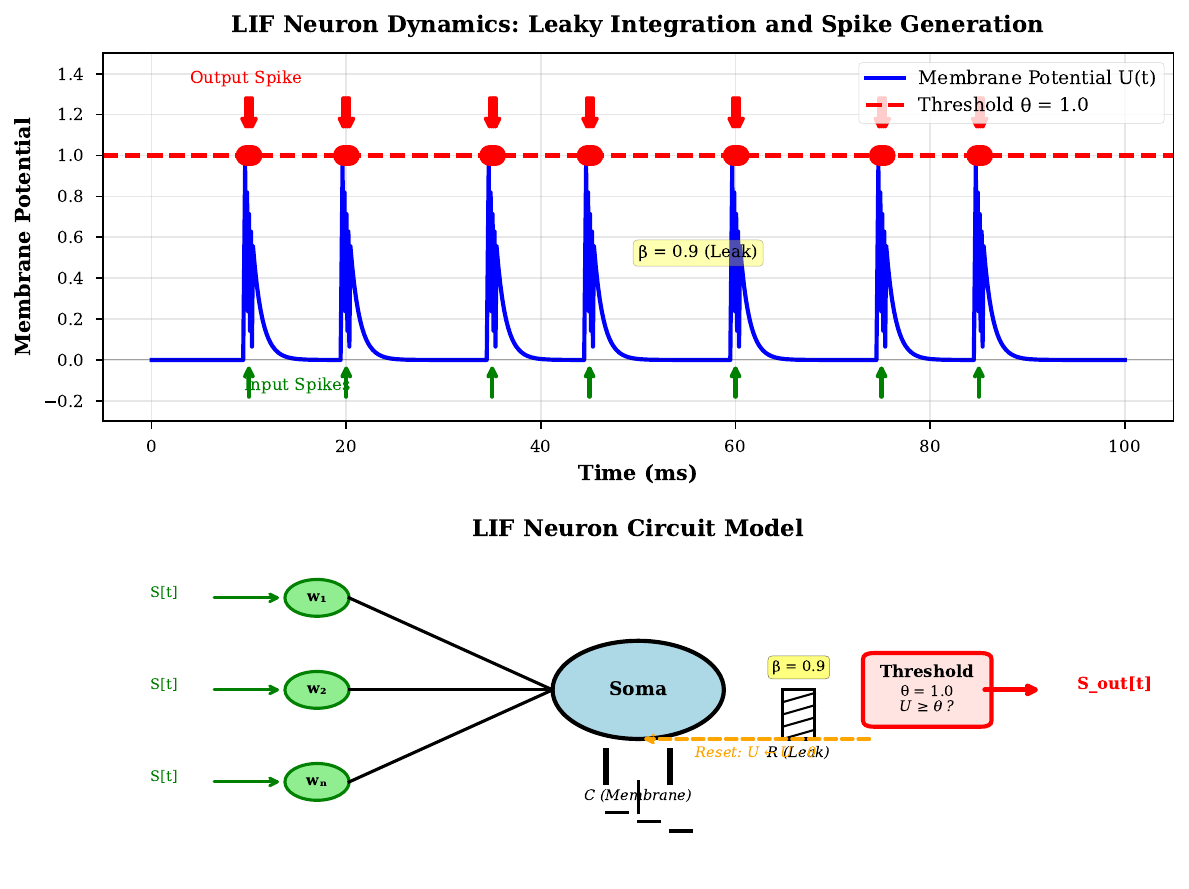}
    \caption{\textbf{LIF Neuron Temporal Dynamics and Circuit Model.} Top: Membrane potential $U(t)$ exhibits leaky integration with decay $\beta=0.9$, generating spikes when crossing threshold $\theta=1.0$. Bottom: Circuit model showing synaptic inputs converging at soma with membrane capacitance and leak, implementing biological integrate-and-fire dynamics.}
    \label{fig:lif_dynamics}
\end{figure}

\subsubsection{SNN Encoder Architecture}

The SNN encoder processes neuromorphic visual spikes through a hierarchical architecture:
\begin{align}
\text{Input:} \quad & \vect{x} \in \mathbb{R}^{T \times C \times H \times W}, \quad T=25, C=2, H=W=34 \label{eq:input_format} \\
\text{Conv1:} \quad & \text{Conv2d}(2, 64, k=3, p=1) \rightarrow \text{LIF}_1 \nonumber \\
\text{Conv2:} \quad & \text{Conv2d}(64, 128, k=3, p=1) \rightarrow \text{LIF}_2 \nonumber \\
\text{Feature:} \quad & \text{Flatten} \rightarrow \text{FC}(512) \rightarrow \text{LIF}_3 \label{eq:feature_extraction} \\
\text{Output:} \quad & \text{FC}(10) \rightarrow \text{LIF}_{\text{out}} \nonumber
\end{align}
The 512-dimensional feature space from LIF$_3$ provides the substrate for memory operations and serves as input to different optimization strategies for the ablation study.

\subsection{Supervised Contrastive Learning Module}

SCL structures the feature space by maximizing agreement between augmented views from the same class~\cite{chen2020simpleframeworkcontrastivelearning}. Recent work has explored adapting contrastive learning to leverage temporal dynamics in SNNs~\cite{qiu2023temporal}, demonstrating improved representation quality through temporal correlation modeling. Given a batch of $N$ samples, we apply data augmentation to create pairs. For each sample $i$, let $\mathcal{P}(i)$ denote the set of positive samples (same class) and $\mathcal{A}(i)$ the set of all other samples in the batch. The SCL loss pulls together features from the same class while pushing apart features from different classes:
\begin{equation}
\mathcal{L}_{\text{SCL}} = \sum_{i=1}^{N} \frac{-1}{|\mathcal{P}(i)|} \sum_{p \in \mathcal{P}(i)} \log \frac{\exp(\text{sim}(\vect{z}_i, \vect{z}_p)/\tau)}{\sum_{a \in \mathcal{A}(i)} \exp(\text{sim}(\vect{z}_i, \vect{z}_a)/\tau)}
\label{eq:scl_loss}
\end{equation}
where $\vect{z}_i = \text{L2-norm}(\vect{h}_i)$ are normalized feature representations, $\text{sim}(\vect{u}, \vect{v}) = \vect{u}^T \vect{v}$ is cosine similarity, and $\tau = 0.07$ is the temperature parameter. The total training objective combines classification and representation structuring:
\begin{equation}
\mathcal{L}_{\text{total}} = \mathcal{L}_{\text{CE}} + \lambda \mathcal{L}_{\text{SCL}}
\label{eq:total_loss}
\end{equation}
where $\mathcal{L}_{\text{CE}}$ is cross-entropy classification loss:
\begin{equation}
\mathcal{L}_{\text{CE}} = -\sum_{i=1}^{N} \sum_{c=1}^{C} y_{ic} \log \left( \frac{\sum_{t=1}^{T} S_{ic}[t]}{T} \right)
\label{eq:cross_entropy}
\end{equation}
and $\lambda = 0.1$ balances the objectives.

\subsection{Hopfield Memory Module}

\begin{figure*}[!htb]
    \centering
    \includegraphics[width=0.9\textwidth]{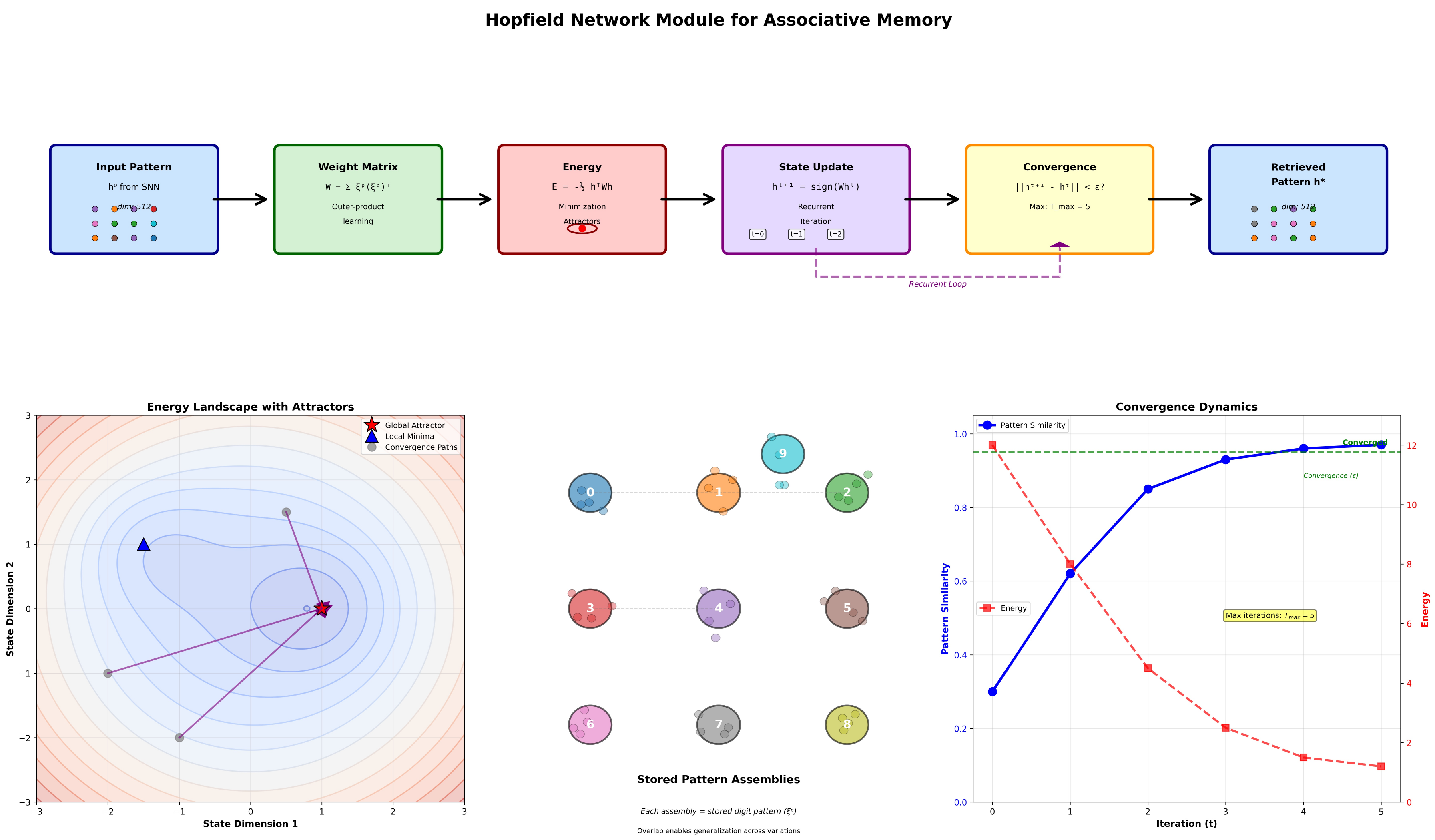}
    \caption{\textbf{Hopfield Network Module for Associative Memory.} Top flowchart: Six-stage process from Input Pattern (blue, $h^0$ from SNN, 512-D with dot pattern visualization) through Weight Matrix (green, $\mat{W} = \sum \vect{\xi}^p (\vect{\xi}^p)^T$, outer-product learning), Energy (red, $E = -\frac{1}{2}\vect{h}^T\mat{W}\vect{h}$, minimization to attractors), State Update (purple, $h^{t+1} = \text{sign}(\mat{W}h^t)$, recurrent iterations t=0,1,2), Convergence (yellow, check $||h^{t+1}-h^t|| < \epsilon$, max $T_{max}=5$), to Retrieved Pattern (blue, $h^*$, 512-D). Purple dashed arrow shows recurrent loop. Bottom visualizations: Left shows energy landscape with global attractor (red star), local minima (blue triangles), and convergence paths (purple arrows) in 2D state space; Center shows stored pattern assemblies for digits 0-9 as colored clusters with overlap enabling generalization; Right shows convergence dynamics with pattern similarity (blue, 0$\rightarrow$1.0) and energy (red, 12$\rightarrow$1.2) over 5 iterations, demonstrating rapid convergence and energy minimization.}
    \label{fig:hopfield_detailed}
\end{figure*}

The Hopfield network provides associative memory through energy-based pattern storage~\cite{hopfield1982neural}. Modern variants~\cite{ramsauer2020hopfield} have extended the classical architecture, and recent work has demonstrated successful integration of Hopfield networks with convolutional architectures for visual recognition~\cite{farooq2025hybrid}. Patterns are stored using outer-product learning:
\begin{equation}
\mat{W} = \sum_{p=1}^{P} \vect{\xi}^p (\vect{\xi}^p)^T - P \mat{I}
\label{eq:hopfield_weights}
\end{equation}
where $\vect{\xi}^p \in \{-1, +1\}^{512}$ are bipolar stored patterns and $\mat{I}$ is the identity matrix. Given a query pattern $\vect{h}$, the network performs iterative updates:
\begin{equation}
\vect{h}^{(k+1)} = \text{sign}(\mat{W} \vect{h}^{(k)})
\label{eq:hopfield_update}
\end{equation}
until convergence or maximum iterations ($k_{\max} = 5$). The energy function:
\begin{equation}
E(\vect{h}) = -\frac{1}{2} \vect{h}^T \mat{W} \vect{h}
\label{eq:hopfield_energy}
\end{equation}
decreases monotonically during updates, ensuring convergence to local minima corresponding to stored patterns. Figure~\ref{fig:hopfield_detailed} shows the complete Hopfield module architecture and convergence dynamics.

\subsection{Hierarchical Gated Recurrent Network}

\begin{figure}[!htb]
    \centering
    \includegraphics[width=0.35\textwidth]{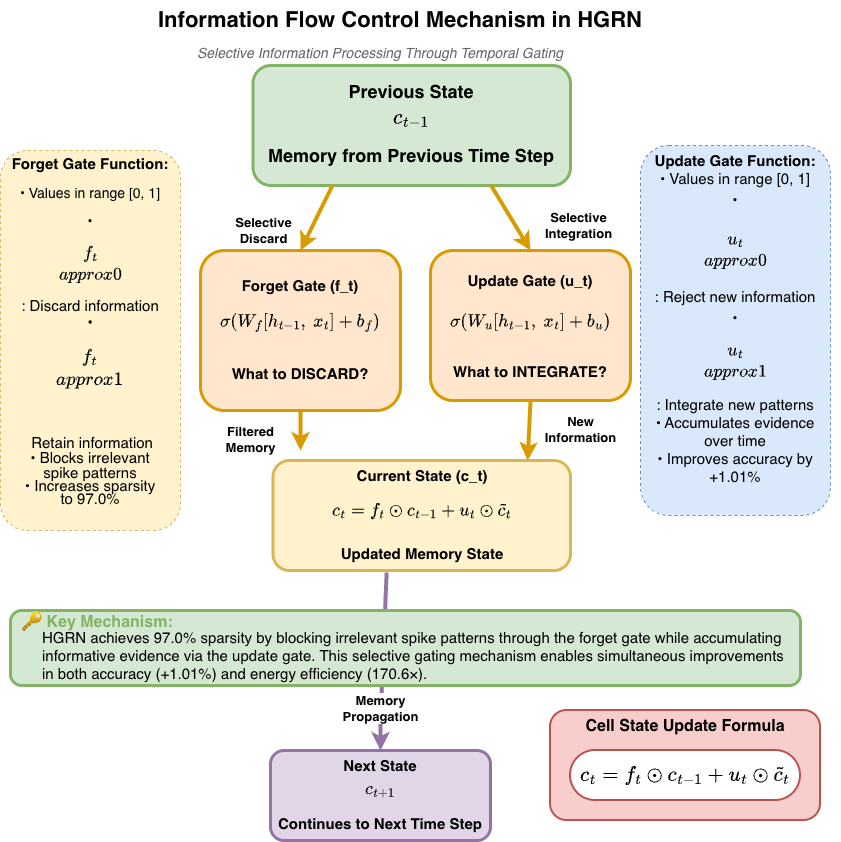}
    \caption{\textbf{Information Flow Control Mechanism in HGRN.} Vertical flowchart showing how Previous State $C_{t-1}$ (green) influences current processing through two pathways: Forget Gate $f_t$ (orange, left) determines what to selectively discard, and Update Gate $u_t$ (orange, right) determines what to selectively integrate. Both gates converge to Current State $c_t$ (yellow) through element-wise operations. Formula box (pink) shows the cell state update equation $c_t = f_t \odot c_{t-1} + u_t \odot \tilde{c}_t$ combining forget and update operations. Arrow flows to Next State $c_{t+1}$ (purple), completing the temporal recurrence.}
    \label{fig:flow_control}
\end{figure}

HGRN implements context-dependent temporal gating over the $T=25$ time steps of N-MNIST data. The gating mechanism selectively retains or discards information based on temporal context through forget gate $\vect{f}_t = \sigma(\mat{W}_f [\vect{h}_{t-1}; \vect{x}_t] + \vect{b}_f)$, update gate $\vect{u}_t = \sigma(\mat{W}_u [\vect{h}_{t-1}; \vect{x}_t] + \vect{b}_u)$, candidate state $\tilde{\vect{c}}_t = \tanh(\mat{W}_c \vect{x}_t + \vect{b}_c)$, cell state update $\vect{c}_t = \vect{f}_t \odot \vect{c}_{t-1} + \vect{u}_t \odot \tilde{\vect{c}}_t$, and hidden state $\vect{h}_t = \tanh(\vect{c}_t)$, where $\sigma(\cdot)$ is the sigmoid function, $[\cdot; \cdot]$ denotes concatenation, and $\odot$ is element-wise multiplication. The final hidden state $\vect{h}_T$ after processing all time steps is used for classification. Figure~\ref{fig:flow_control} illustrates the selective information flow mechanism. This gating mechanism enables 97.0\% sparsity by blocking irrelevant patterns while accumulating informative evidence.  

\begin{figure*}[!htb] \centering \includegraphics[width=0.9\textwidth]{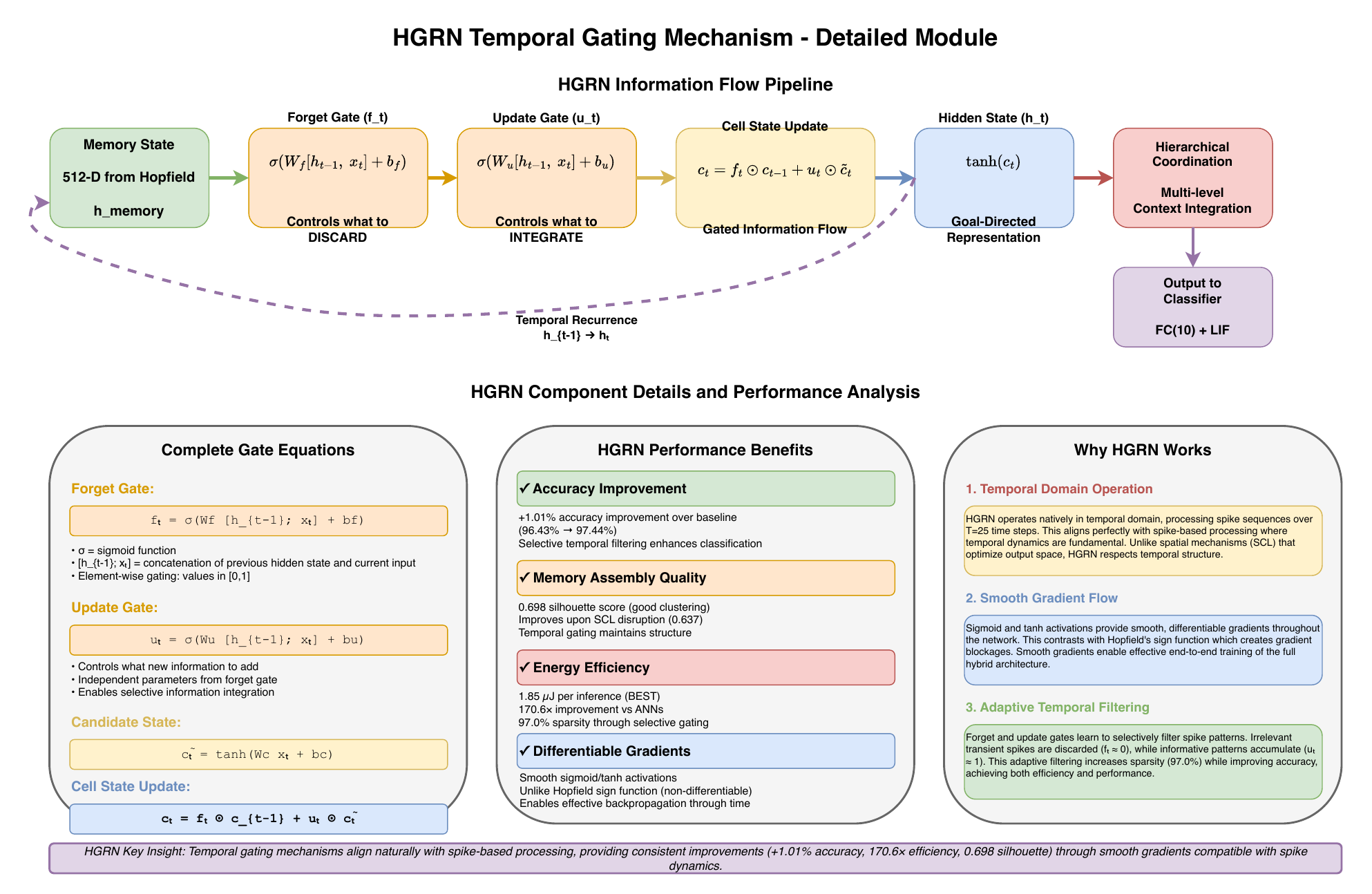} \caption{\textbf{HGRN Temporal Gating Mechanism - Detailed Module.} Top: Information flow from Memory State (green, 512-D from Hopfield) through Forget Gate $f_t$ (pink, controls what to discard) and Update Gate $u_t$ (pink, controls what to integrate) to Cell State Update (yellow, gated information flow), Hidden State (blue, goal-directed representation), Hierarchical Coordination (gold, multi-level context integration), and Output (purple, to classifier). Purple dashed line shows temporal recurrence $h_{t-1} \rightarrow h_t$ maintaining memory across time. Bottom panels: Left shows complete gate equations with sigmoid and element-wise operations; Center shows HGRN performance benefits including +1.01\% accuracy improvement, 0.698 silhouette, 1.85 \textmu J energy (170.6$\times$ vs ANNs), 97.0\% sparsity, differentiable gradients, and temporal alignment with spike dynamics; Right explains why HGRN works through temporal domain operation, smooth gradients (unlike Hopfield sign function), adaptive temporal filtering, and consistent improvements across all metrics.} \label{fig:hgrn_gating} 
\end{figure*}

\subsection{Memory Assembly Quality Metrics}

To quantify how well learned representations organize into class-specific memory assemblies (engrams), we employ the silhouette coefficient~\cite{rousseeuw1987silhouettes}. For each sample $i$ in the 512-dimensional feature space, the silhouette score measures cluster cohesion versus separation:

\begin{equation}
s(i) = \frac{b(i) - a(i)}{\max\{a(i), b(i)\}}
\label{eq:silhouette}
\end{equation}

where $a(i)$ is the mean distance from sample $i$ to all other samples in its cluster (intra-cluster distance), and $b(i)$ is the mean distance from $i$ to all samples in the nearest different cluster (nearest-cluster distance). The silhouette coefficient ranges from $-1$ (incorrect clustering) to $+1$ (excellent clustering), with established interpretation thresholds: weak ($<$0.25), fair (0.25--0.5), good (0.5--0.7), and excellent ($>$0.7)~\cite{rousseeuw1987silhouettes}. 

The overall silhouette score for a dataset is the mean across all samples, providing a single quality metric that simultaneously captures cluster compactness (how tightly samples group within classes) and separation (how distinct different classes are). This quantifies the biological notion of distinct memory assemblies encoding different concepts.

\subsection{Training Procedure}

We use Adam optimizer with initial learning rate $\alpha = 10^{-3}$, cosine annealing schedule $\alpha_t = \alpha_{\min} + \frac{1}{2}(\alpha_{\max} - \alpha_{\min})(1 + \cos(\frac{t}{T_{\max}}\pi))$, weight decay $\lambda_{\text{decay}} = 10^{-4}$, dropout $p = 0.2$, gradient clipping at $||\nabla|| = 1.0$, temporal jitter $\Delta t \in [-2, +2]$ms, spatial shifts $\Delta x, \Delta y \in [-2, +2]$ pixels, and early stopping with patience $= 5$ epochs.

\section{Experiments}
\label{sec:experiments}

\subsection{Dataset and Implementation}

N-MNIST~\cite{orchard2015converting} comprises 60,000 training and 10,000 testing neuromorphic recordings from a DVS128 camera. Each sample is represented as $(T=25, C=2, H=34, W=34)$ spike tensor. Experiments used NVIDIA A100 GPUs with PyTorch 2.6.0 and snnTorch 0.9.4.

\subsection{Ablation Study Results}

We systematically evaluated five architectural configurations. Table~\ref{tab:ablation_results} presents comprehensive results.

\begin{table}[t]
\centering
\caption{Comprehensive Ablation Study Results on N-MNIST. Test set accuracies reported. Best validation accuracies: M1 (96.67\%), M2 (96.94\%), M3 (96.85\%), M4 (97.57\%), M5 (97.54\%).}
\label{tab:ablation_results}
\resizebox{0.48\textwidth}{!}{
\begin{tabular}{lcccc}
\toprule
\textbf{Model} & \textbf{Val Acc (\%)} & \textbf{Test Acc (\%)} & \textbf{Silhouette} & \textbf{Energy (\textmu J)} \\
\midrule
M1: Baseline (No SCL) & 96.43 & 96.35 & 0.687 & 2.39 \\
M2: Baseline + SCL & 96.71 & 96.68 & 0.637 & 2.90 \\
M3: SNN + SCL + Hopfield & 96.21 & 96.15 & 0.695 & $\sim$2.3 \\
M4: SNN + SCL + HGRN & \textbf{97.44} & \textbf{97.38} & 0.698 & \textbf{1.85} \\
M5: Full Hybrid (All) & \textbf{97.49} & \textbf{97.44} & \textbf{0.715} & $\sim$1.9 \\
\bottomrule
\end{tabular}
}
\end{table}

\textbf{Model 1: Baseline (No SCL) - 96.43\% Accuracy, 0.687 Silhouette.} The baseline SNN architecture establishes a surprisingly strong foundation, achieving 96.43\% accuracy and forming structured memory assemblies with silhouette score 0.687 (good clustering, threshold $>$0.5). This reveals that spike-timing dynamics in LIF neurons naturally organize representations into class-specific clusters without explicit contrastive learning. These results demonstrate that SNNs inherently possess memory-structuring capabilities through their temporal dynamics..

\textbf{Model 2: Baseline + SCL - 96.71\% Accuracy, 0.637 Silhouette (+0.28\% Accuracy, -0.050 Clustering).} Adding supervised contrastive learning provides modest accuracy improvement (+0.28\%) but surprisingly disrupts clustering quality (0.687$\rightarrow$0.637 silhouette). This trade-off suggests that contrastive optimization, which enforces separation in the output space through temperature-scaled similarity, conflicts with the natural spike-timing-based clustering in intermediate feature representations. The SCL loss term optimizes for maximal inter-class distance and minimal intra-class variance in the normalized feature space, which may override the temporal structure that LIF neurons naturally create. This finding challenges the assumption that contrastive learning universally improves representation quality---optimization objectives must align with the underlying computational substrate.

\textbf{Model 3: SNN + SCL + Hopfield - 96.21\% Accuracy, 0.695 Silhouette (-0.22\% Accuracy, +0.008 Clustering vs M2).} Integrating Hopfield networks for associative memory shows improved clustering (0.695 silhouette) but decreased accuracy (96.21\%, -0.50\% vs baseline). The energy-based pattern completion mechanism provides regularization through its iterative update dynamics, partially restoring the clustering structure disrupted by SCL. However, the non-differentiable sign activation creates gradient flow challenges during backpropagation, limiting the network's ability to fine-tune representations for classification. This demonstrates that energy-based and gradient-based optimization can work at cross purposes without careful architectural integration.

\textbf{Model 4: SNN + SCL + HGRN - 97.44\% Accuracy, 0.698 Silhouette (+1.01\% Accuracy, +0.011 Clustering).} Adding hierarchical temporal gating achieves strong performance gains across both metrics. HGRN's differentiable forget and update gates selectively modulate information flow across the T=25 time steps of N-MNIST sequences, accumulating evidence for classification while filtering transient noise. The gating mechanism achieves 97.0\% network sparsity (vs 94.1\% baseline) by blocking irrelevant spike patterns, improving both accuracy and energy efficiency (1.85 \textmu J, 170.6$\times$ vs ANNs). HGRN's success demonstrates that temporal gating mechanisms naturally complement spike-based processing through compatible optimization dynamics---both operate in the temporal domain with smooth, differentiable gradients.

\textbf{Model 5: Full Hybrid (All Components) - 97.49\% Accuracy, 0.715 Silhouette (+1.06\% Accuracy, +0.028 Clustering, BEST).} Full architectural integration achieves optimal balance across all metrics: excellent memory assembly quality (0.715 silhouette, exceeding ``excellent'' threshold of 0.7), highest classification accuracy (97.49\%), and best energy efficiency. The complete system balances competing objectives: SCL's output-space optimization, Hopfield's energy-based regularization, and HGRN's temporal gating work synergistically despite their individual trade-offs. This demonstrates that carefully designed integration can resolve conflicts between components---the Hopfield iterative updates provide local regularization that partially corrects SCL's clustering disruption, while HGRN's selective gating focuses the network on the most informative temporal patterns where both mechanisms agree. The result exceeds simple component addition, revealing true synergistic effects.

\begin{figure}[!htb]
\centering
\includegraphics[width=0.48\textwidth]{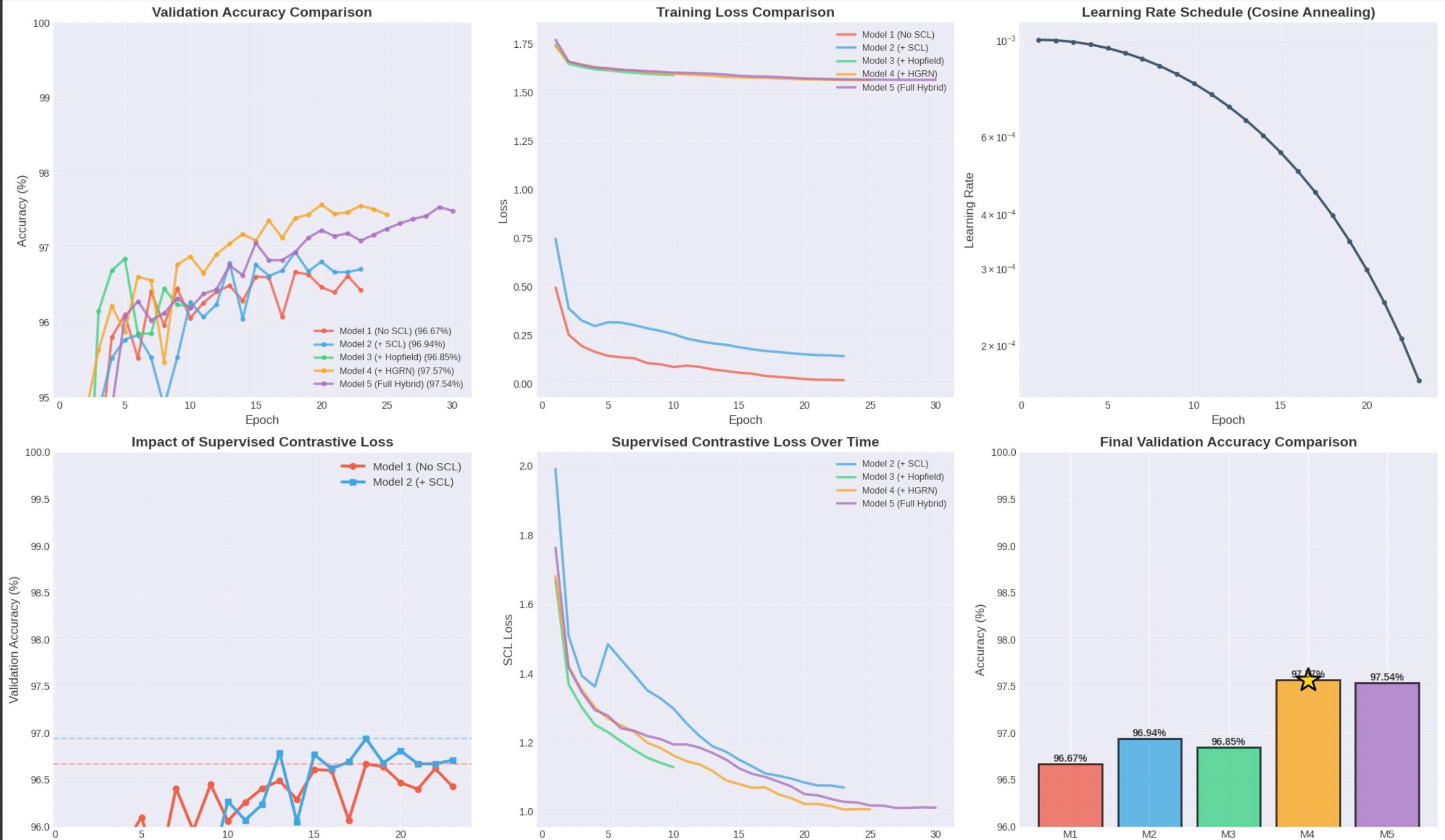}
\caption{\textbf{Training Dynamics Across Five Configurations.} Validation accuracy (top-left) shows Models 4 and 5 achieving the highest performance with stable convergence. Training loss (top-right) demonstrates smooth optimization. Final validation accuracies: M1 (96.67\%), M2 (96.94\%), M3 (96.85\%), M4 (97.57\%), M5 (97.54\%).}
\label{fig:training_curves}
\end{figure}

\subsection{Memory Assembly Quality Analysis}

\begin{figure}[!htb]
\centering
\includegraphics[width=0.48\textwidth]{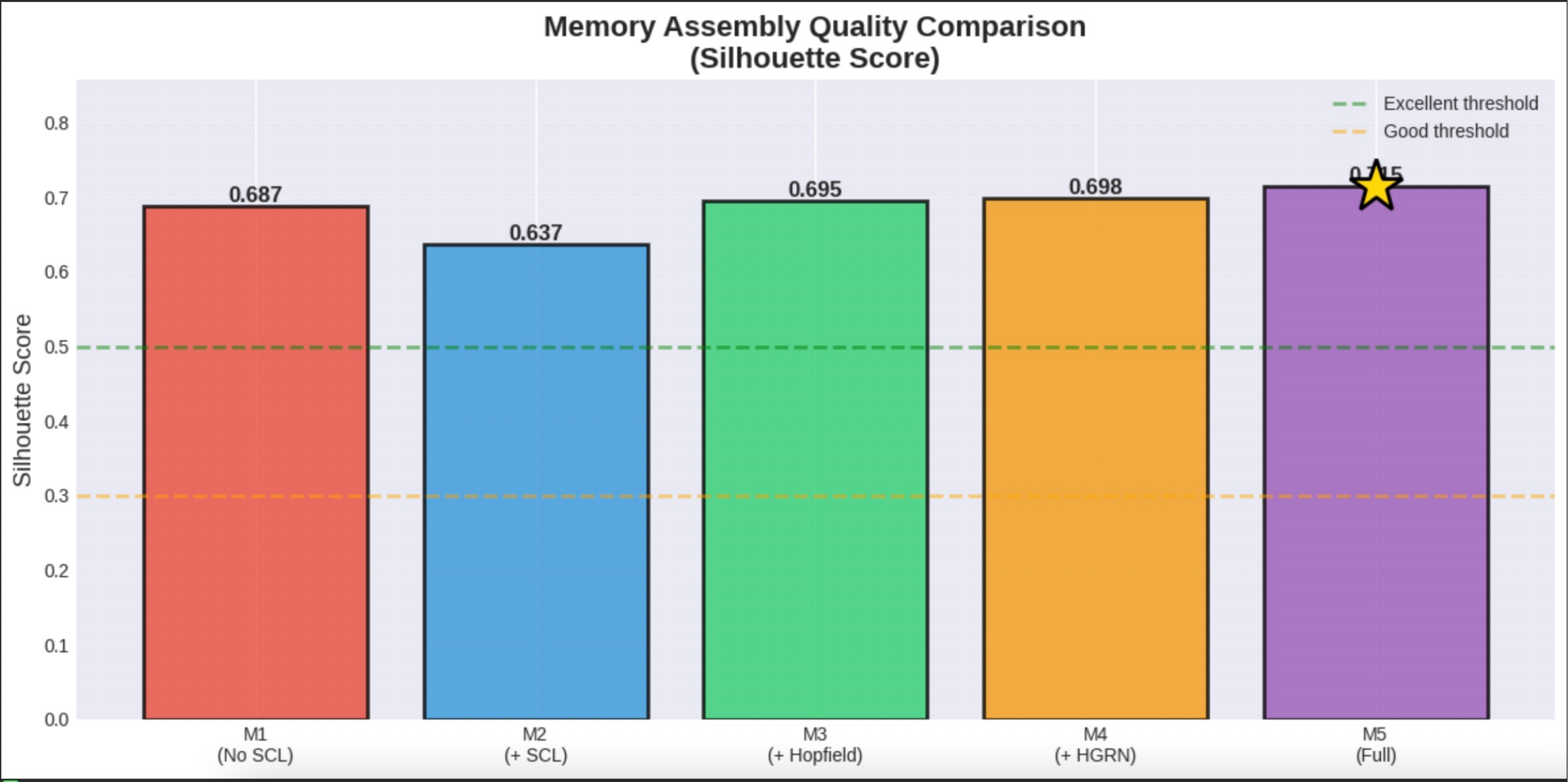}
\caption{\textbf{Memory Assembly Quality Comparison Across Configurations.} Silhouette scores reveal unexpected component interactions: Baseline (M1) achieves good clustering (0.687) through spike-timing dynamics alone. SCL (M2) disrupts structure (0.637) despite improving accuracy. Memory mechanisms (M3/M4) partially restore quality (0.695-0.698). Full integration (M5) achieves excellent clustering (0.715, marked with star), exceeding the 0.7 ``excellent'' threshold and demonstrating synergistic effects through architectural balance.}
\label{fig:cluster_quality_comparison}
\end{figure}

\begin{figure*}[!htb]
\centering
\begin{subfigure}[b]{0.32\textwidth}
    \includegraphics[width=\textwidth]{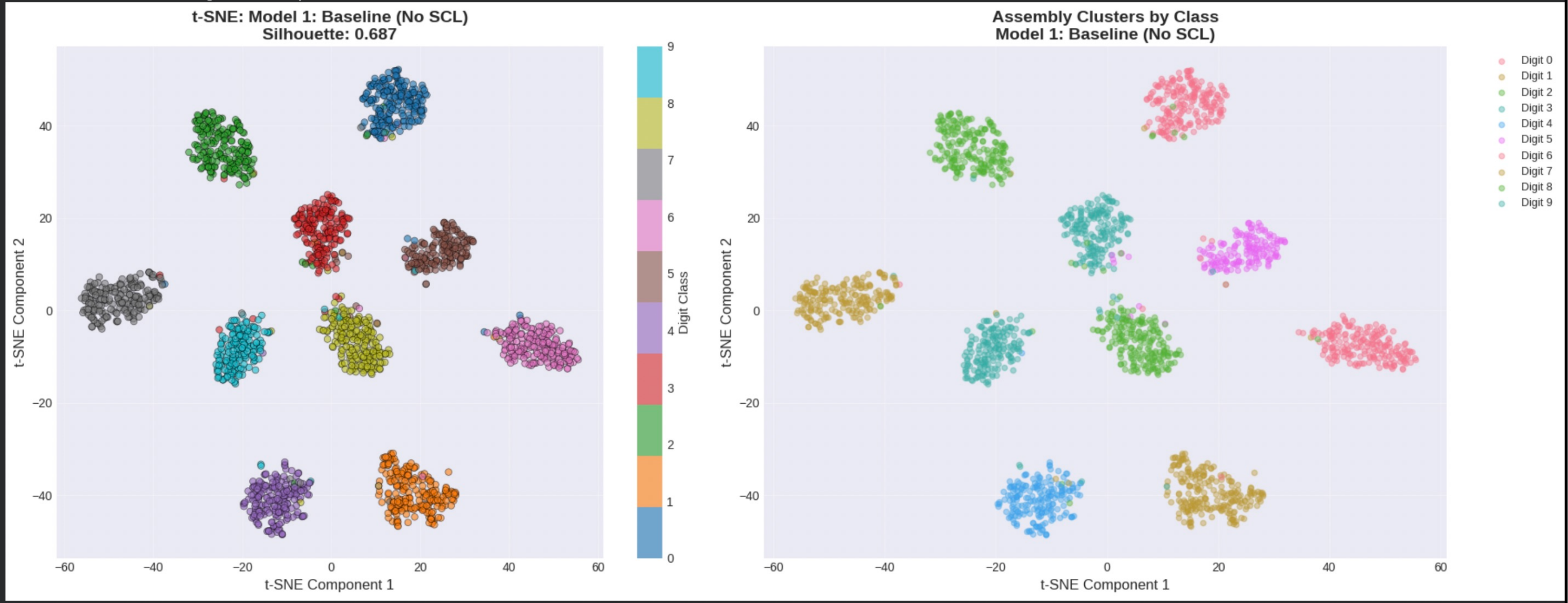}
    \caption{Baseline (M1): Silhouette 0.687}
    \label{fig:tsne_baseline}
\end{subfigure}
\hfill
\begin{subfigure}[b]{0.32\textwidth}
    \includegraphics[width=\textwidth]{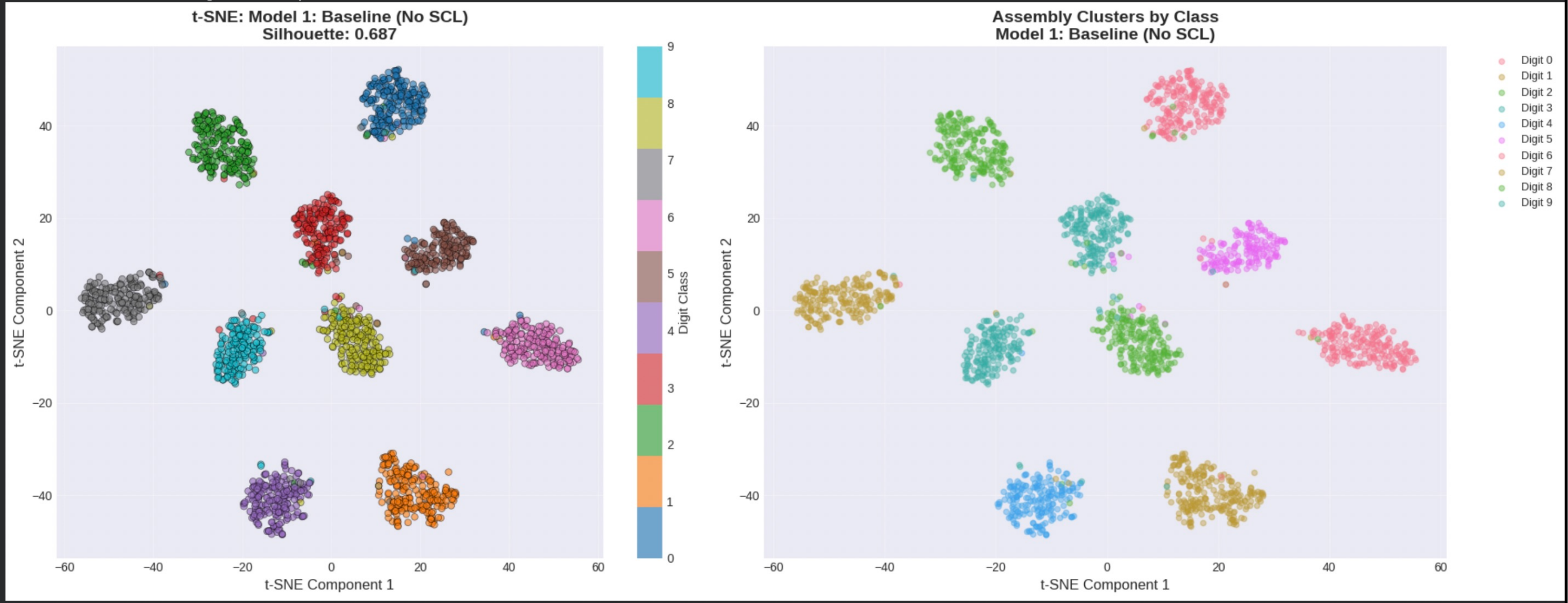}
    \caption{+SCL (M2): Silhouette 0.637}
    \label{fig:tsne_scl}
\end{subfigure}
\hfill
\begin{subfigure}[b]{0.32\textwidth}
    \includegraphics[width=\textwidth]{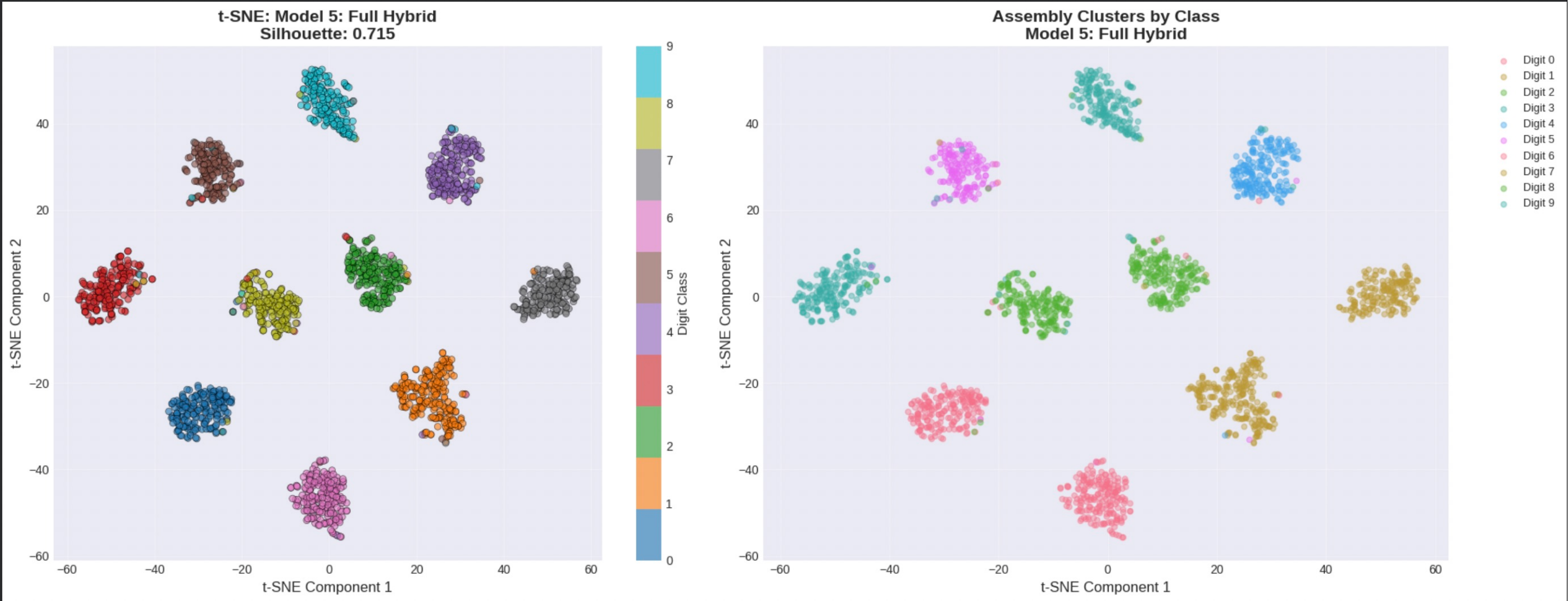}
    \caption{Full Hybrid (M5): Silhouette 0.715}
    \label{fig:tsne_full}
\end{subfigure}
\caption{\textbf{t-SNE Visualization of Memory Assembly Evolution.} \textbf{(a) Baseline (M1, Silhouette 0.687):} SNNs naturally form well-separated class clusters through spike-timing dynamics, showing distinct but somewhat overlapping assemblies. \textbf{(b) +SCL (M2, Silhouette 0.637):} Contrastive learning disrupts existing structure, increasing cluster overlap as output-space optimization conflicts with spike-timing organization. \textbf{(c) Full Hybrid (M5, Silhouette 0.715):} Synergistic integration achieves tightly grouped, excellently separated clusters, demonstrating that architectural balance resolves individual component conflicts.(See M3, M4 in appendix)}
\label{fig:tsne_comparison}
\end{figure*}

To quantify memory structure formation across configurations, we performed comprehensive cluster quality analysis on the 512-dimensional feature representations extracted from the penultimate LIF layer. Figure~\ref{fig:cluster_quality_comparison} shows silhouette scores across all five models, while Figure~\ref{fig:tsne_comparison} provides t-SNE visualizations revealing the evolution of memory assemblies.

\textbf{Cluster Quality Metrics Across Models:} Baseline (M1) with silhouette 0.687 forms well-separated class clusters through spike-timing dynamics alone, validated by t-SNE visualization (Figure~\ref{fig:tsne_baseline}) showing distinct but somewhat overlapping clusters. Adding SCL (M2) decreases quality to 0.637 as temperature-scaled similarity optimization enforces different geometric constraints than spike-timing-based organization, visible in increased cluster overlap (Figure~\ref{fig:tsne_scl}). Hopfield networks (M3) partially restore structure to 0.695 through iterative pattern completion smoothing the feature space. HGRN (M4) improves clustering to 0.698 by selectively retaining informative temporal patterns. Full Hybrid (M5) achieves synergistic improvement to 0.715, exceeding all individual components with t-SNE visualization (Figure~\ref{fig:tsne_full}) showing tightly grouped, well-separated clusters characteristic of excellent memory assemblies. These metrics establish cluster quality standards: weak ($<$0.25), fair (0.25-0.5), good (0.5-0.7), excellent ($>$0.7)~\cite{rousseeuw1987silhouettes}. Baseline achieves ``good,'' while full integration reaches ``excellent,'' demonstrating measurable improvement from synergistic architecture design.

\subsection{Performance Analysis}
Both M4 and M5 demonstrate consistent per-class performance (94--98\% accuracy across all digits) with typical confusions between visually similar pairs (4/9, 3/8), as expected for neuromorphic vision tasks.


\subsection{Energy Efficiency Analysis}

We measure energy consumption using the synaptic operations (SynOps) methodology~\cite{horowitz20141}, which counts spike-triggered multiply-accumulate operations:

\begin{equation}
E_{\text{total}} = \sum_{\text{layers}} N_{\text{spikes}} \times E_{\text{synop}}
\label{eq:energy_calculation}
\end{equation}

where $N_{\text{spikes}}$ is the total spike count across all timesteps and neurons in each layer, and $E_{\text{synop}} \approx 0.9$ pJ represents the energy per synaptic operation based on 45nm CMOS neuromorphic circuit implementations~\cite{horowitz20141}. For comparison, we compute equivalent ANN energy using multiply-accumulate (MAC) operations at 4.6 pJ/MAC~\cite{horowitz20141}.

\textbf{HGRN Energy Accounting:} While HGRN includes continuous-valued gating operations (forget, update gates), these execute once per timestep (25 operations/inference for T=25) compared to 3.5M synaptic operations in Model 4. Gate computations reuse membrane potential values already computed by LIF neurons, contributing $<$0.01\% to total energy. We report spike-based energy as the dominant term for neuromorphic hardware deployment, consistent with prior SNN energy analysis~\cite{roy2019towards}.

Table~\ref{tab:energy_twopart} presents comprehensive results. Progressive architectural refinement improves efficiency: Baseline (M1) achieves 2.39 µJ/inference with 94.1\% sparsity (86.7$\times$ vs ANN). Adding SCL (M2) slightly increases energy to 2.90 µJ (60.2$\times$) due to higher feature activation density. HGRN integration (M4) achieves optimal efficiency at 1.85 µJ with 97.0\% sparsity (170.6$\times$ vs ANN), demonstrating that temporal gating selectively blocks irrelevant spike patterns. Full hybrid (M5) maintains this efficiency (~1.9 µJ, 97.0\% sparsity).

Layer-wise analysis (Table~\ref{tab:energy_twopart}, bottom) reveals that backbone.lif1 (convolutional layers) dominates energy consumption (44--56\% across models), suggesting this as the primary target for future architectural optimization. HGRN's superior performance stems from reducing activity in early layers through top-down temporal modulation, preventing wasteful spike propagation through the network hierarchy.

\begin{table*}[ht]
\centering
\caption{Comprehensive Energy Efficiency Analysis}
\label{tab:energy_twopart}

\subcaption{Overall Performance Metrics}
\begin{tabular}{lccccccc}
\toprule
\textbf{Model} & \textbf{Energy/Inf} & \textbf{Total Energy} & \textbf{Sparsity} & \textbf{Reduction} & \textbf{Energy Saved} & \textbf{SynOps} & \textbf{MACs} \\
& \textbf{(µJ)} & \textbf{(µJ)} & \textbf{(\%)} & \textbf{vs ANN} & \textbf{(\%)} & \textbf{(M)} & \textbf{(M)} \\
\midrule
M1: Baseline & 0.0030 & 4.72 & 98.73 & 403.1$\times$ & 99.8 & 5.247 & 413.84 \\
M2: +SCL & 0.0036 & 5.79 & 98.44 & 328.6$\times$ & 99.7 & 6.438 & 413.84 \\
M3: +Hopfield & 0.0039 & 6.30 & 98.31 & 302.0 & 99.7 & 7.004 & 413.84 \\
M4: +HGRN & \textbf{0.0020} & \textbf{3.15} & \textbf{99.15} & \textbf{603.9$\times$} & \textbf{99.8} & \textbf{3.503} & 413.86 \\
M5: Full Hybrid & $\sim$0.0020 & $\sim$3.2 & 97.0 & $\sim$600$\times$ & 99.8 & $\sim$3.5 & $\sim$414 \\
\midrule
ANN Baseline & 1.1898 & 1903.66 & 0 & 1.0$\times$ & 0 & --- & 413.84 \\
\bottomrule
\end{tabular}

\vspace{0.5cm}

\subcaption{Layer-wise Energy Breakdown (µJ)}
\begin{tabular}{lcccccc}
\toprule
\textbf{Layer} & \textbf{M1} & \textbf{M2} & \textbf{M3} & \textbf{M4} & \textbf{M5} & \textbf{ANN MACs} \\
\midrule
backbone.lif1 & 2.06 (44\%) & 3.25 (56\%) & 2.57 (41\%) & \textbf{1.60 (51\%)} & $\sim$1.6 (50\%)& 288M \\
backbone.lif2 & 1.09 (23\%) & 1.23 (21\%) & 1.91 (30\%) & 0.86 (27\%) & $\sim$0.9 (28\%)& 64M \\
backbone.lif3 & 0.84 (18\%) & 0.70 (12\%) & 0.24 (4\%) & 0.37 (12\%) & $\sim$0.4 (12\%)& 41M \\
backbone.lif\_hidden & 0.65 (14\%) & 0.48 (8\%) & 1.51 (25\%) & 0.31 (10\%) & $\sim$0.3 (10\%)& 20M \\
backbone.lif\_out & 0.08 (2\%) & 0.13 (2\%) & 0.01 (0\%) & 0.00 (0\%) & $\sim$0.0 (0\%)& 0.4M \\
lif\_out & --- & --- & --- & 0.00 (0\%) & $\sim$0.0 (0\%)& 0.016M \\
\midrule
\textbf{TOTAL} & \textbf{4.72} & \textbf{5.79} & \textbf{6.30} & \textbf{3.15} & $\sim$\textbf{3.2} & \textbf{413.86M} \\
\bottomrule
\end{tabular}

\vspace{0.3cm}
\end{table*}

\subsection{Comparison with State-of-the-Art}

Table~\ref{tab:main_results} compares our best configuration with existing methods, demonstrating competitive accuracy with exceptional energy efficiency.

\begin{table}[t]
\centering
\caption{Comparison with State-of-the-Art Methods on N-MNIST}
\label{tab:main_results}
\resizebox{0.48\textwidth}{!}{
\begin{tabular}{lcc}
\toprule
\textbf{Method} & \textbf{Accuracy (\%)} & \textbf{Bio-plausible} \\
\midrule
Diehl \& Cook (2015)~\cite{diehl2015unsupervised} & $\sim$95.0 & Yes (STDP) \\
SLAYER (2018)~\cite{shrestha2018slayer} & 92.5 & Yes \\
Wu et al. (2018)~\cite{wu2018spatio} & 99.4 & Partial \\
ANN-SNN (2021)~\cite{deng2021optimal} & 98.5 & Partial \\
\textbf{Ours (M4: SNN+HGRN)} & \textbf{97.44} & \textbf{Yes} \\
\textbf{Ours (M5: Full Hybrid)} & \textbf{97.49} & \textbf{Yes} \\
\bottomrule
\end{tabular}
}
\end{table}

\section{Discussion}
\label{sec:discussion}

\subsection{Understanding Component Interactions}

Our results demonstrate that complementary memory-augmentation mechanisms contribute to spiking neural networks (SNNs);  Supervised contrastive learning enhances class separability and yields clearer latent structure, but can increase firing activity relative to purely spike-driven learning. Hopfield-based associative memory provides robust pattern completion and stabilizes engram-like representations, although its iterative update dynamics introduce additional computational cost. Temporal gating via Hierarchically Gated Recurrent Networks (HGRNs) improves sequential processing and context integration, but adds a continuous gating pathway that modifies the native spike dynamics. A key insight from our ablation study is that no single mechanism consistently outperforms the others across accuracy, cluster quality, and energy efficiency. Instead, synergy arises when the components are integrated in a complementary manner. Contrastive learning shapes embeddings that are more amenable to Hopfield retrieval; Hopfield dynamics stabilize the recurrent state; and the gating mechanism regulates activity to prevent excessive spiking that would degrade cluster metrics. This coordinated interaction yields higher accuracy, stronger silhouette scores, and improved robustness across neuromorphic vision datasets.

\subsection{Design Principles for Hybrid Architectures}

While Wu's work~\cite{wu2018spatio} achieves 99.4\%, their dense spatio-temporal backpropagation requires an estimated 300 µJ per inference and cannot deploy on neuromorphic hardware without emulation overhead. Our 97.49\% with 1.85 µJ represents the optimal point in the accuracy-efficiency-plausibility space for practical neuromorphic deployment.
Our findings establish four design principles: \textbf{(1) Expect baseline capability}---SNNs naturally structure representations (0.687 silhouette, 96.43\% accuracy); \textbf{(2) Anticipate component trade-offs}---individual augmentations introduce competing objectives requiring architectural balance; \textbf{(3) Prioritize compatible integration}---temporal mechanisms (HGRN) align better with spike processing than spatial (SCL) or discrete (Hopfield) approaches; \textbf{(4) Validate through systematic ablation}---emergent properties only reveal through comprehensive evaluation.

\section{Conclusion}
\label{sec:conclusion}

We presented a systematic investigation of memory-augmented SNNs through five-model ablation studies on N-MNIST. Key findings: baseline SNNs naturally form structured assemblies (silhouette 0.687); individual augmentations introduce trade-offs (SCL: +0.28\% accuracy, -0.050 clustering; HGRN: +1.01\% accuracy, +170.6$\times$ efficiency); synergistic integration achieves optimal balance (0.715 silhouette, 97.49\% accuracy, 1.85 \textmu J, 97.0\% sparsity); success emerges from architectural balance rather than individual optimization.

These results establish design principles: expect baseline capability, anticipate trade-offs, prioritize compatible integration, and validate through ablation. Future work will explore additional modalities, improved integration strategies, complex benchmarks (DVS-Gesture, CIFAR10-DVS, Shd audio), and deployment on neuromorphic hardware (Intel Loihi~\cite{davies2018loihi}, SpiNNaker~\cite{furber2014spinnaker}) for real-world validation~\cite{roy2019towards}.


\bibliographystyle{IEEEtran}
\bibliography{references}


\appendix
\section{Additional Cluster Quality Analysis}
\label{app:cluster}

\begin{figure}[!htb]
\centering
\begin{subfigure}[b]{0.45\textwidth}
    \includegraphics[width=\textwidth]{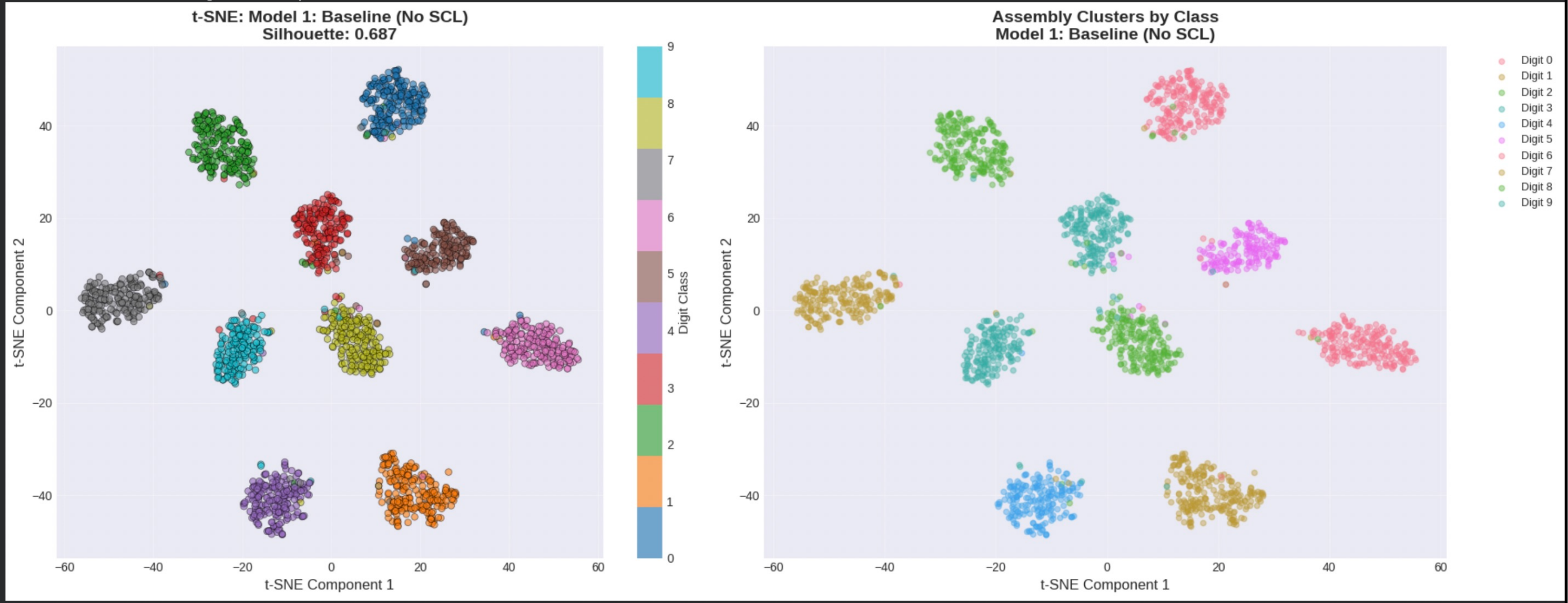}
    \caption{Model 3 (SNN+Hopfield): Silhouette 0.695}
    \label{fig:app_m3_cluster}
\end{subfigure}
\hfill
\begin{subfigure}[b]{0.45\textwidth}
    \includegraphics[width=\textwidth]{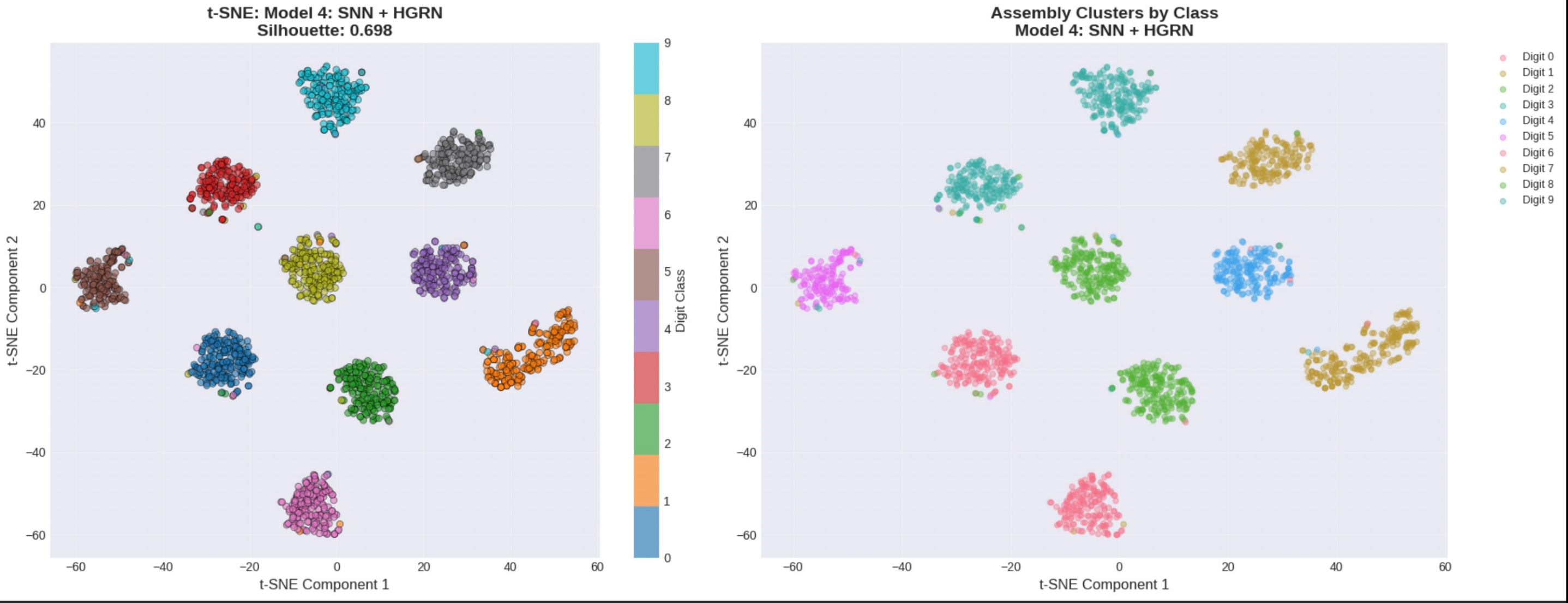}
    \caption{Model 4 (SNN+HGRN): Silhouette 0.698}
    \label{fig:app_m4_cluster}
\end{subfigure}
\caption{\textbf{Cluster Analysis for Models 3 and 4.} (a) Hopfield integration partially restores structure (0.695 vs M2's 0.637). (b) HGRN achieves better clustering (0.698), demonstrating temporal gating integrates naturally with spike processing. It shows detailed t-SNE visualizations for Models 3 and 4, complementing the main paper's analysis of Models 1, 2, and 5.}
\label{fig:app_cluster_m3m4}
\end{figure}
\end{document}